%% file: paper.tex
\documentclass[conference]{IEEEtran}
\usepackage{times}

\usepackage[numbers]{natbib}
\usepackage{multicol}
\usepackage{graphicx}
\usepackage{amsmath}
\usepackage{amssymb}
\usepackage{booktabs}
\usepackage{makecell}
\usepackage{float}
\usepackage{xcolor}
\usepackage[bookmarks=true]{hyperref}

\newif\ifanonymous
\anonymousfalse

\begin{document}

\title{WCM: World-Cognition Model for Generalizable Human--Robot Interaction}

\ifanonymous
\author{Author Names Omitted for Anonymous Review. Paper-ID [add your ID here]}
\else
\author{%
\IEEEauthorblockN{Yuzhen Chen\textsuperscript{1}, KC Zhou\textsuperscript{2}}
\IEEEauthorblockA{%
\textsuperscript{1}Harvard University\quad
\textsuperscript{2}Massachusetts Institute of Technology}%
}
\fi

\maketitle

\begin{abstract}
\input{sections/abstract}

\end{abstract}

\IEEEpeerreviewmaketitle

\input{sections/introduction}

\input{sections/method}

\input{sections/experiments}

\input{sections/conclusion}

\ifanonymous
\section*{Acknowledgments}
\else
\section*{Acknowledgments}
We thank T. Lihe and Andrew Hoang for implementation support. This work was
funded by Cyberbrain, Inc.; the system, datasets, and related IP are owned by
Cyberbrain, Inc.
\fi


\clearpage
\bibliographystyle{plainnat}
\bibliography{references}

\input{sections/appendix}

\end{document}

%% file: sections/abstract.tex
Language agents can now interact fluently with users in software, but robots still struggle to bring comparable interaction to physical tasks. Current robot-control paradigms, including vision-language-action policies and world-model-based planners, are mainly optimized for instruction execution, leaving users with little visibility into why an action is chosen and few mechanisms to redirect, correct, or teach the robot through interaction.
To solve this problem, we present the World-Cognition Model (WCM), a human-centered embodied agent built on the SLAK architecture (Sensing, Logic, Action, and Knowledge) and an asynchronous runtime. SLAK separates perception, reasoning, control, and memory, while the runtime allows reasoning, dialogue, and execution to proceed concurrently. WCM further introduces a human-in-the-loop teaching mode that enables users to interactively teach the robot difficult or long-horizon tasks. Teaching episodes and autonomous task rollouts are refined into chain-of-thought supervision to continually improve the model.
WCM achieves a 73.8\% average success rate across nine real-world human--robot interaction tasks, including tasks held out from CoT fine-tuning and a long-horizon task learned through teaching.

%% file: sections/introduction.tex
\section{Introduction}

Language has become a common interface for robot control, but following language instructions is not the same as human--robot interaction. In software, language agents allow users to refine goals, ask for explanations, and steer a task as it unfolds. In embodied systems, however, language is still often treated as an input condition for action generation rather than a persistent channel for interaction, explanation, and feedback-driven learning. As a result, even robots that can execute language instructions offer limited support for mid-execution redirection, human-readable explanations, or learning from user feedback.

Vision-language-action (VLA) and robot foundation models
~\cite{rt1,rt2,openvla,octo,pi0,pi05,grootn1}
and recent world-model systems for physical AI, robot imagination, and
manipulation~\cite{worldmodels,daydreamer,robodreamer,genie,cosmos,dreamgen,zhou2026gem}
expose this gap in different ways. VLA policies can accept language commands, but language usually serves as a conditioning signal for action prediction rather than a persistent channel through which a user can intervene, revise the goal, or teach the robot during execution. Their decisions are also difficult to inspect: even when a pretrained vision-language backbone provides semantic understanding, the final motion is produced by an action decoder or policy, leaving the link between high-level reasoning and physical behavior implicit. World-model-based planners face a related limitation. They reason through predicted future states or imagined rollouts, which can support planning, but these rollouts are not by themselves human-readable explanations, nor do they naturally provide a mechanism for real-time conversational correction. Thus, both families advance language-conditioned execution, but neither makes interaction, legibility, and feedback-driven learning a first-class part of the control loop.

Interactive robots therefore require more than a stronger
language-conditioned policy: perception, reasoning, action, memory, and human
feedback must remain connected throughout execution. The robot should ground a
user's instruction in the current scene, expose the reason behind its next
action, revise that action when the user intervenes, and turn interaction itself
into future training signal~\cite{clarkbrennan,horvitz}.

We instantiate this view in the World-Cognition Model (WCM), a human-centered
embodied agent built on the SLAK architecture and an asynchronous runtime. SLAK
separates Sensing, Logic, Action, and Knowledge, making perception, reasoning,
control, and memory explicit rather than hidden inside a single opaque policy.
The runtime allows reasoning, dialogue, and execution to proceed concurrently,
while WCM converts autonomous rollouts and human teaching episodes into
chain-of-thought supervision for continual improvement~\cite{cot,distill}.

We evaluate WCM on a low-cost functional mobile-manipulation platform and
propose nine real-world human--robot interaction tasks covering object
retrieval, handover, tool use, drawer manipulation, and trash disposal. WCM
achieves a 73.8\% average success rate, including tasks held out from CoT
fine-tuning and a long-horizon task learned through teaching.

Our contributions are threefold. 
\textbf{First,} we introduce WCM, an interactive embodied system with explicit
SLAK layers and asynchronous execution. 
\textbf{Second,} we propose a human-in-the-loop teaching and CoT distillation
pipeline that turns autonomous rollouts and teaching episodes into training
signal. 
\textbf{Third,} we build a low-cost mobile-manipulation platform and introduce a
nine-task real-world HRI suite for evaluating language-guided interaction,
real-time correction, and task learning.

%% file: sections/method.tex
\section{Method: The World-Cognition Model}
\label{sec:method}

\subsection{System Architecture and Runtime}
\label{sec:system}

\subsubsection{SLAK Architecture}
\label{sec:slak}
\begin{figure*}[t]
  \centering
  \includegraphics[width=\textwidth]{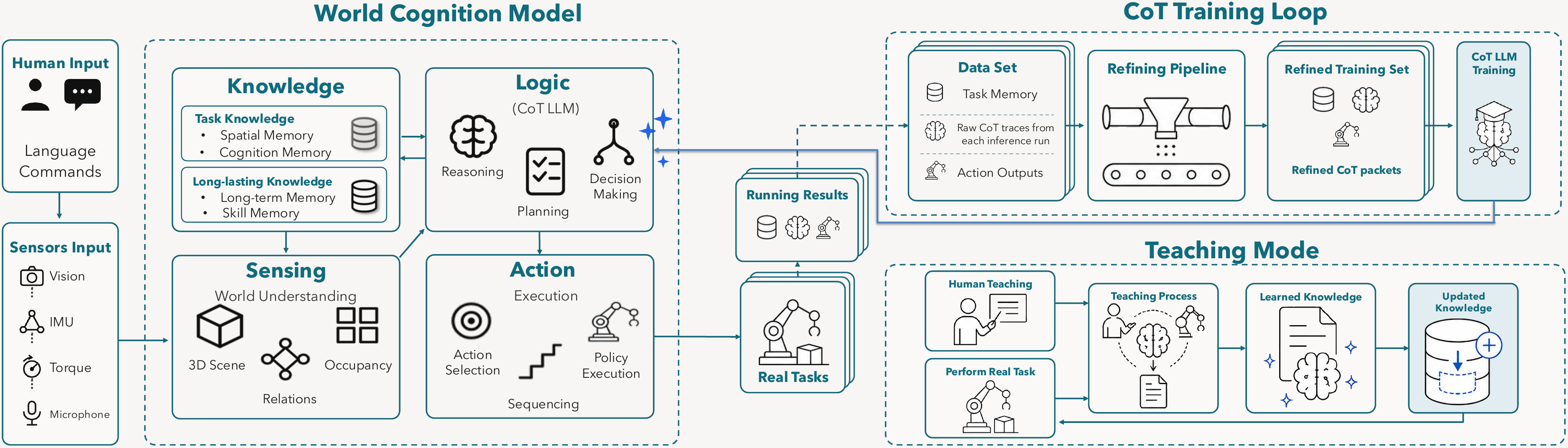}
\caption{\textbf{The SLAK architecture.}
WCM organizes sensing, logic, action, and knowledge as explicit layers connected
to a CoT distillation pipeline. Autonomous executions and human teaching
episodes are refined into CoT training data for continual improvement.}
  \label{fig:slak}
\end{figure*}
SLAK is the core system architecture of WCM, consisting of four explicit layers:
Sensing, Logic, Action, and Knowledge (Fig.~\ref{fig:slak}). These layers
separate scene understanding, reasoning, robot control, and memory while
communicating through explicit state exchange. This design lets WCM update the
scene, goal, memory, and action outcomes during execution, rather than hiding
the full perception-to-action process inside a single end-to-end policy
\cite{rt2,openvla,pi0}.

\begin{figure*}[t]
  \centering
  \includegraphics[width=\textwidth]{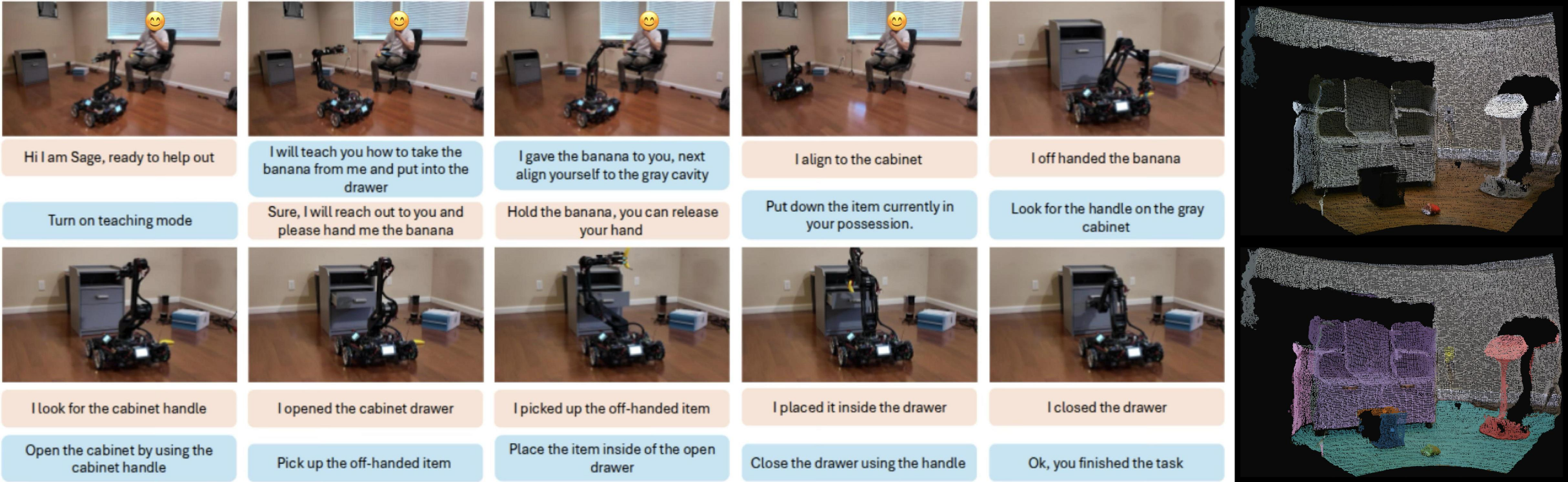}
 \caption{\textbf{Human-in-the-loop teaching and sensing output.}
Left: natural-language teaching of a long-horizon task. Right: raw RGB-D
observations and the corresponding interaction-aware scene state. The episode is
refined and distilled with autonomous rollouts.}
  \label{fig:teaching}
\end{figure*}

\paragraph{Sensing layer}
The Sensing layer converts multimodal sensor streams into an
interaction-aware scene state. The right side of Fig.~\ref{fig:teaching}
visualizes this conversion from raw RGB-D observations to a structured
scene state with semantic and part-level interaction cues. It integrates vision, depth, audio, IMU, and
force/torque signals to maintain local 3D structure, object relations,
occupancy, and semantic cues around the robot. Unlike streaming 3D perception
methods that primarily reconstruct geometry from image sequences
\cite{stream3d,page4d}, or open-vocabulary grounding and segmentation models
that localize language-specified concepts as boxes or masks
\cite{groundingdino,groundedsam,sam3}, WCM estimates the scene properties needed
for robot interaction. These include pixel- and part-level cues for grasping,
opening, placement, and disposal, as well as coarse physical and material
attributes such as rigidity, fragility, and grasp suitability
\cite{matpredict,ewen2024feel}. The resulting state informs Logic about available local
interactions and the physical constraints that should shape the next action.



\paragraph{Logic layer}
The Logic layer is WCM's reasoning, planning, and decision-making module. It
conditions each decision on the user's current instruction, the structured scene
state from Sensing, and the relevant memory retrieved from Knowledge. 

Rather than committing to a complete task script upfront, Logic follows an
incremental plan--reason--validate loop. At step $t$, Logic forms a decision
context $x_t=(s_t,u_t,m_t)$, where $s_t$ denotes the structured scene state,
$u_t$ the user's current instruction or goal, and $m_t$ the relevant task memory
retrieved from Knowledge. Given this context, Logic produces a reasoning trace,
the next action, and its expected outcome:
\begin{equation}
  (r_t,a_t,\hat{o}_t)=f_{\theta}(x_t), \qquad
  v_t =
  \begin{cases}
  1, & \Delta(o_t,\hat{o}_t)\le\epsilon,\\
  0, & \text{otherwise}.
  \end{cases}
  \label{eq:logic-validate}
\end{equation}
Here, $f_{\theta}$ is the Logic model, $\theta$ denotes its parameters, $o_t$ is
the realized outcome returned after execution, $\Delta$ measures the mismatch
between expected and realized outcomes, and $\epsilon$ is a validation threshold.
If $v_t=1$, the task proceeds; otherwise Logic re-reasons from the updated state
or asks the user for clarification.

This stepwise structure keeps
decisions grounded in the current scene, exposes the robot's intent and expected
outcome in natural language, and limits error propagation when the environment
changes or the user redirects the task
\cite{saycan,palme,huang2022inner}.

\paragraph{Knowledge layer}
The Knowledge layer stores and updates the state and experience WCM accumulates
during task execution and long-term use. It maintains short-term task memory,
including the current goal, observed objects, spatial context, recent actions,
outcomes, and user feedback, as well as longer-term knowledge such as task
experience, object-interaction history, and skill memory. This memory allows
WCM to preserve context across a continuous interaction rather than treating each
decision as an isolated response to the current observation.

\paragraph{Action layer}
The Action layer converts Logic's selected step into executable robot behavior.
Given a step and its expected outcome, Action binds it to the appropriate
low-level skill or policy, sequences the required motion commands, and executes
them on the robot hardware. These skills cover navigation, reaching, grasping,
placing, handover, and simple articulated-object interactions, and are reused
across tasks rather than trained as task-specific policies. During execution,
Action reports status, failures, and realized outcomes back to Logic and
Knowledge for validation and replanning when necessary. This separation keeps
high-level reasoning independent from the underlying controller, allowing the
action library or hardware stack to be improved without changing the reasoning
interface \cite{liang2023code,sutton1999between}.


During execution, Sensing, Logic, and Action write scene changes, active goals,
reasoning states, action outcomes, failures, and user feedback into Knowledge as
a shared task state. Logic retrieves this state when deciding the next step,
keeping execution, interaction feedback, and later learning connected within the
same system loop.

Taken together, SLAK makes WCM's execution state explicit across sensing,
reasoning, action, and memory. This explicit state interface provides the basis
for the asynchronous runtime, which coordinates concurrent reasoning, execution,
and state updates while preserving a shared task context.

\subsubsection{Asynchronous Runtime}
\label{sec:async}

Building on SLAK's explicit state interface, the asynchronous runtime decouples
reasoning, execution, and state updates during task execution. In a conventional
robot-control pipeline, perception, reasoning, action, and feedback are processed
sequentially, so model inference can block the robot from acting or delay its
response to new user commands. WCM instead allows the reasoner to propose
actions, the executor to carry out committed actions on the hardware, and the
state updater to incorporate new observations, outcomes, and user feedback
without forcing these processes to run in a single blocking sequence.

This asynchronous design does not mean that WCM blindly follows an outdated
action queue. Before committing queued actions and after execution, the runtime
applies the validation check in Eq.~\eqref{eq:logic-validate} using the latest
scene state, user input, and realized outcome. If the context no longer matches
the expectation, stale actions are discarded and Logic re-reasons from the
updated state. This decoupling preserves responsiveness in dynamic environments,
allowing the robot to move, explain, listen for corrections, and update the task
plan as the situation changes.

\subsection{CoT Distillation from Rollouts and Teaching }
\label{sec:learning}

To connect execution with learning, WCM converts autonomous rollouts and
human-guided teaching episodes into step-level CoT supervision. For an episode
of length $T$, each decision record stores the scene, instruction, memory,
reasoning, selected action, expected outcome, and realized outcome:
\begin{equation}
  e = \{d_t\}_{t=1}^{T}, \quad
  d_t = (s_t, u_t, m_t, r_t, a_t, \hat{o}_t, o_t),
  \label{eq:decision-record}
\end{equation}
where $s_t$ is the scene state, $u_t$ the user instruction, $m_t$ the retrieved
memory, $r_t$ the Logic reasoning, $a_t$ the selected action, $\hat{o}_t$ the
expected outcome, and $o_t$ the realized outcome.

These records come from two sources: autonomous executions, including both
successes and failures, and human-guided teaching episodes for difficult or
long-horizon tasks. Let $\mathcal{D}_{\mathrm{auto}}$ and
$\mathcal{D}_{\mathrm{teach}}$ denote the corresponding decision records. The offline refinement pipeline produces refined CoT examples
$\mathcal{D}'_{\mathrm{auto}}=\mathcal{R}(\mathcal{D}_{\mathrm{auto}})$ and
$\mathcal{D}'_{\mathrm{teach}}=\mathcal{R}(\mathcal{D}_{\mathrm{teach}})$,
where $\mathcal{R}$ denotes the refinement procedure.
Because teaching episodes contain direct human guidance and clearer task
decomposition, we weight them more heavily during distillation. Let
$\mathcal{L}_{\mathrm{auto}}(\theta)$ and
$\mathcal{L}_{\mathrm{teach}}(\theta)$ denote the CoT negative log-likelihood
losses over $\mathcal{D}'_{\mathrm{auto}}$ and
$\mathcal{D}'_{\mathrm{teach}}$, respectively. The weighted distillation
objective is
\begin{equation}
  \mathcal{L}_{\mathrm{distill}}(\theta)
  =
  \mathcal{L}_{\mathrm{auto}}(\theta)
  + \lambda \mathcal{L}_{\mathrm{teach}}(\theta),
  \label{eq:distill}
\end{equation}
where $\lambda \ge 1$ controls the relative weight of human-guided teaching
examples. WCM minimizes this objective to update the reasoning model.

\subsubsection{Autonomous Data Flywheel}
\label{sec:data-flywheel}

During autonomous execution, WCM records the reasoning-action process behind
each completed or failed task. Successful episodes provide positive evidence for
effective task decomposition, grounding, action selection, and outcome
validation, but their raw traces may still contain redundant reasoning, indirect
steps, or unnecessary recovery attempts. The refinement pipeline therefore turns
successful executions into cleaner and more direct CoT traces, making the
underlying logic easier for the model to imitate. Failed episodes are also
useful, but not as positive demonstrations directly. Instead, they expose where
the model misunderstood the scene, selected an ineffective action, predicted the
wrong outcome, or repeated an unnecessary recovery step. These failure records
contribute in two ways: they are stored in Knowledge as reusable experience for
future reasoning, and they are sent to the offline refinement pipeline, where
the failed reasoning path is converted into a more targeted corrected CoT trace
for distillation. Unlike VLA-style training on teleoperated state-action
trajectories~\cite{rt2,openvla,pi0}, WCM uses the decision records in
Eq.~\eqref{eq:decision-record} as reasoning supervision rather than
motor-action supervision.

Raw execution records are therefore not used directly for training. Real robot
runs can contain redundant reasoning, repeated attempts, indirect action choices,
or ineffective intermediate actions, even when the final task succeeds. A
stronger model ensemble reviews the recorded reasoning and outcomes, removes
unnecessary detours, corrects inconsistent decision paths when needed, and
produces cleaner reasoning traces through voting or consensus. The result
preserves the causal structure of real execution while turning noisy robot
behavior into higher-quality CoT supervision
\cite{cot,nye2021show,lightman2024verify,uesato2022solving}.

The refined autonomous examples are distilled under the shared objective in
Eq.~\eqref{eq:distill}, closing the autonomous data flywheel~\cite{robocat,robogen}: execution produces decision records, refinement
improves the supervision, and distillation improves later executions.
Over time, WCM accumulates reusable reasoning patterns for object interaction,
action sequencing, outcome prediction, and failure recovery, rather than
memorizing a single task script.

\subsubsection{Human-in-the-Loop Teaching Mode}
\label{sec:humanloop}

The autonomous data flywheel improves WCM from its own successes and failures,
but some unseen or long-horizon tasks remain inefficient to discover through
trial and error alone. For such cases, WCM introduces a human-in-the-loop
teaching mode. Rather than requiring low-level teleoperation, the user teaches
through natural language, guiding the robot step by step by indicating what to
attend to, which object or part to act on, and when to advance to the next stage.



During teaching, WCM records the same step-level decision format as autonomous
execution, augmented with explicit human guidance and feedback. These examples
are often more informative than unguided rollouts because they provide clearer
task decomposition, direct correction signals, and human choices about how a
difficult task should be staged. This makes them especially useful for
long-horizon or previously unseen tasks, where autonomous exploration alone may
be slow or brittle.

Teaching episodes enter the same refinement pipeline and shared distillation
objective as autonomous execution records, but are weighted by $\lambda$ in
Eq.~\eqref{eq:distill} because they provide direct human guidance. In the short
term, taught procedures are stored in the Knowledge layer for reuse on similar
tasks. In the long term, refined teaching examples are distilled into the
reasoning model, allowing the robot to internalize human guidance as reusable
task competence~\cite{cot,distill,deng2023implicit,deng2024explicit}.

%% file: sections/experiments.tex
\section{Experiments}
\label{sec:experiments}

We evaluate WCM on a low-cost mobile-manipulation platform and a nine-task
real-world HRI suite, measuring task success, held-out transfer, teaching-based
skill acquisition, system efficiency, and ablated components.
\subsection{Hardware}
\label{sec:exp-hw}

WCM runs on a low-cost mobile-manipulation platform: a Mecanum-wheel base, a
5-DOF arm ($\sim$1.3\,m reach), and a $640\times480$ RGB-D camera. The complete
robot costs under \$2{,}000, while reasoning runs off-board on a shared
NVIDIA RTX~5090 over the local network; we therefore report robot hardware cost
separately from compute.

\subsection{Nine-Task Human--Robot Interaction Suite}
\label{sec:exp-hri}

We propose a nine-task real-world human--robot interaction suite to evaluate WCM
beyond isolated manipulation primitives. Each task begins with a spoken
instruction and requires language understanding, scene grounding, and physical
execution. The suite covers object retrieval, handover, tool use, drawer
manipulation, and trash disposal. Per-task success rates, trials, and run times
are reported in Table~\ref{tab:interaction}, with frame-by-frame demonstrations
in Figs.~\ref{fig:demo-popcorn}--\ref{fig:demo-bottletrash}
(Appendix~\ref{app:demos}).

The robot communicates in real time during these tasks: it listens, confirms,
asks, and explains while acting. Four tasks are held out from CoT fine-tuning yet
still succeed, demonstrating transfer beyond the training tasks. The hardest
task, ``Take the screwdriver to the drawer,'' cannot be solved reliably before
teaching; after interactive teaching it reaches $69\%$ success, and after
distillation it rises to $82\%$, testing the full teaching-to-distillation loop.

\subsection{Case-Study Comparison}
\label{sec:exp-cmp}

We present a case-study comparison with X-Square's WALL-OSS, reproducing its
task suite as closely as our platform allows. Because the robots differ in body,
kinematics, sensors, and execution stack, this comparison is illustrative rather
than a controlled benchmark. Per-task speed, GPU occupancy, and success rates
appear in Appendix~\ref{app:benchmark} (Table~\ref{tab:benchmark}).

\subsection{Ablations}
\label{sec:exp-abl}

Finally, we ablate the asynchronous runtime and the interaction-aware Sensing
layer on a shared five-task subset. Removing asynchrony forces the robot to wait
between reasoning and action, increasing end-to-end run time by $1.4$--$1.7\times$.
Replacing Sensing with a plain VLM and 2D grounding sharply reduces success on
part-level interaction tasks; Screwdriver-to-Drawer falls from $69\%$ to $0\%$.
Results are reported in Appendix~\ref{app:ablation} (Table~\ref{tab:ablation}).

%% file: sections/conclusion.tex
\section{Conclusion}
\label{sec:conclusion}

We present WCM, a human-centered embodied agent that combines SLAK,
asynchronous execution, and CoT distillation from rollouts and teaching. WCM
maintains task context, explains intent, responds to mid-execution corrections,
and improves from interaction. On a low-cost mobile-manipulation platform, WCM
achieves a 73.8\% average success rate across nine real-world tasks, including
held-out tasks and a long-horizon task learned through teaching. While currently
limited by hardware and a case-study evaluation, WCM points toward robots that
can act, explain, be corrected, and improve with people in the loop.

%% file: sections/appendix.tex
\clearpage
\onecolumn
\appendices

\section{Human--Robot Interaction Demos}
\label{app:demos}
\input{wcm_assets/tables/table2_interaction.tex}

\begin{figure}[H]
  \centering
  \includegraphics[width=\linewidth]{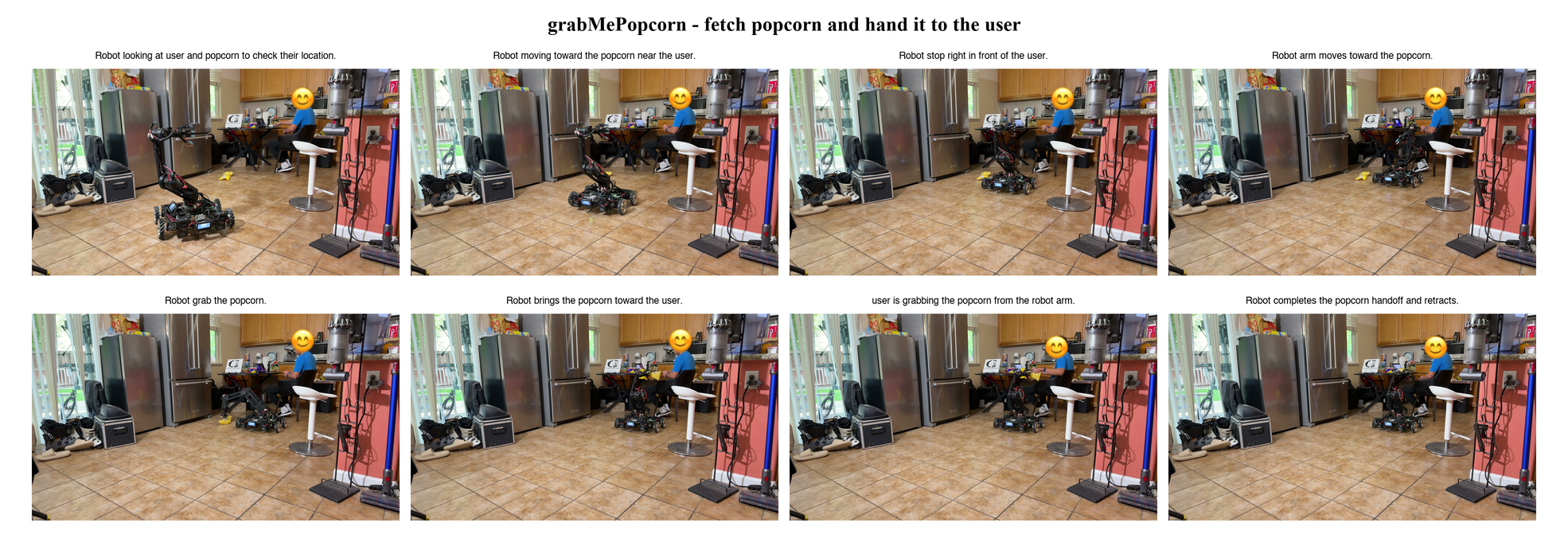}
  \caption{\textbf{``Grab me the popcorn.''} Object retrieval and handover.}
  \label{fig:demo-popcorn}
\end{figure}

\begin{figure}[H]
  \centering
  \includegraphics[width=\linewidth]{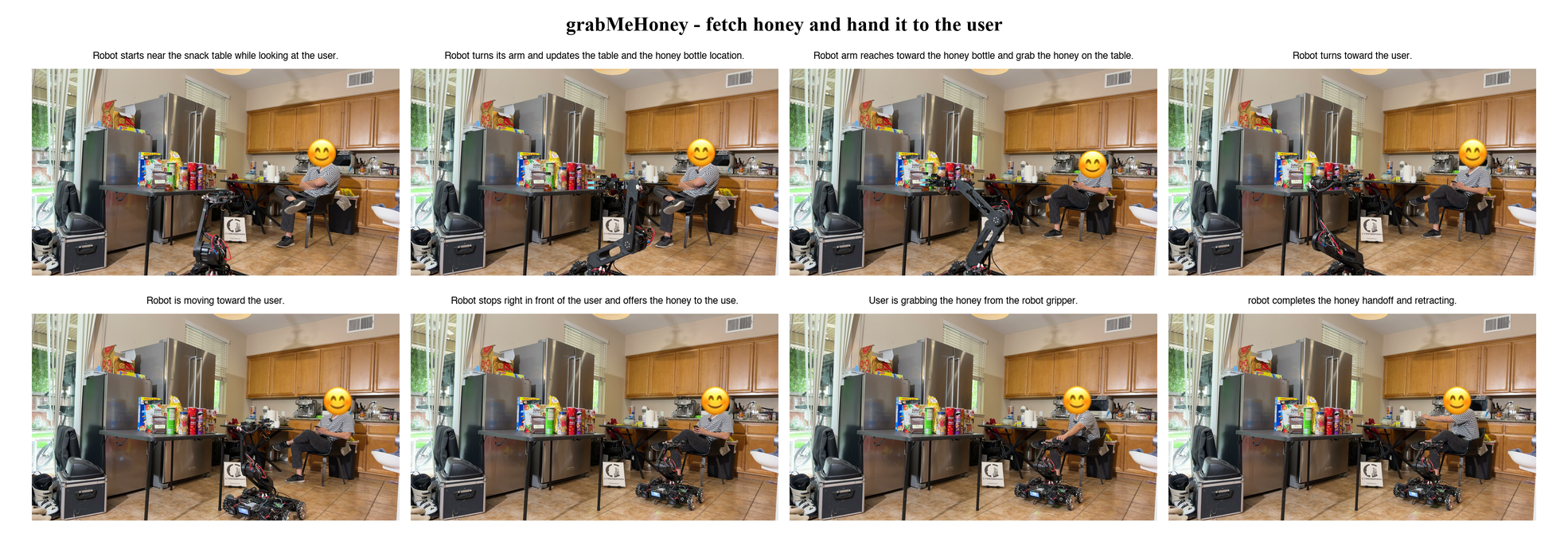}
  \caption{\textbf{``Hand me the honey.''} The robot locates the honey, grasps it, and hands it to the person.}
  \label{fig:demo-honey}
\end{figure}

\begin{figure}[H]
  \centering
  \includegraphics[width=\linewidth]{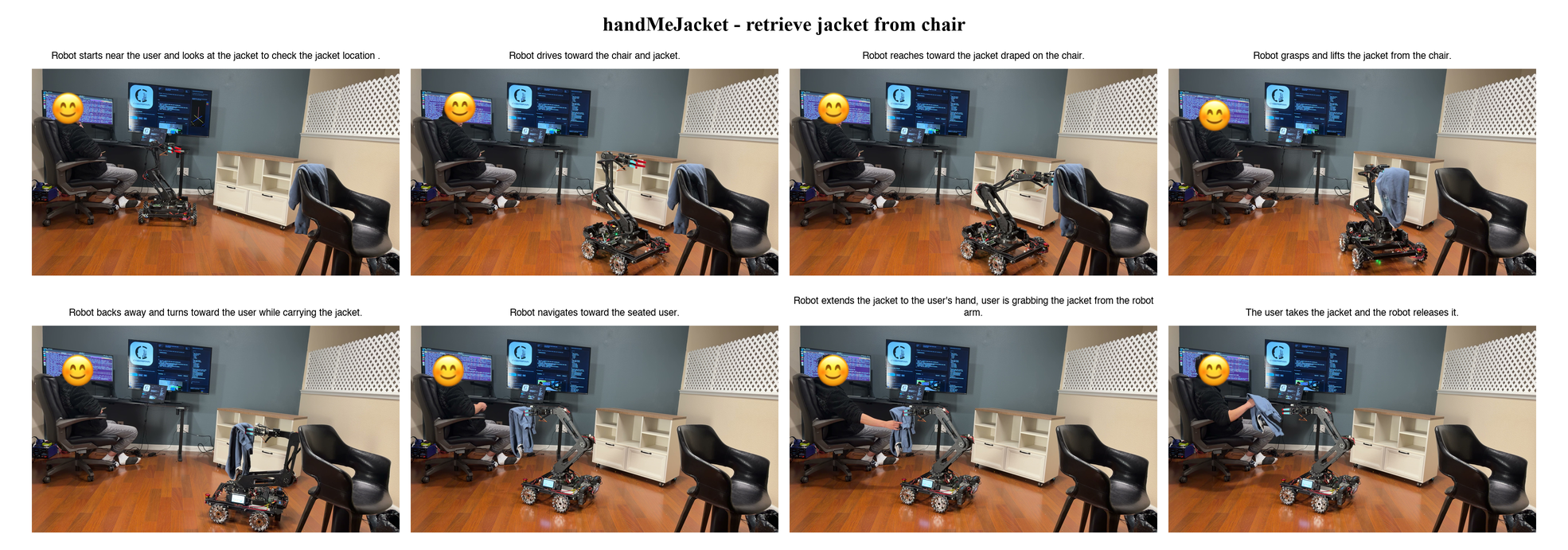}
  \caption{\textbf{``Hand me the jacket.''} Grasping a deformable object and delivering it to the person.}
  \label{fig:demo-jacket}
\end{figure}

\begin{figure}[H]
  \centering
  \includegraphics[width=\linewidth]{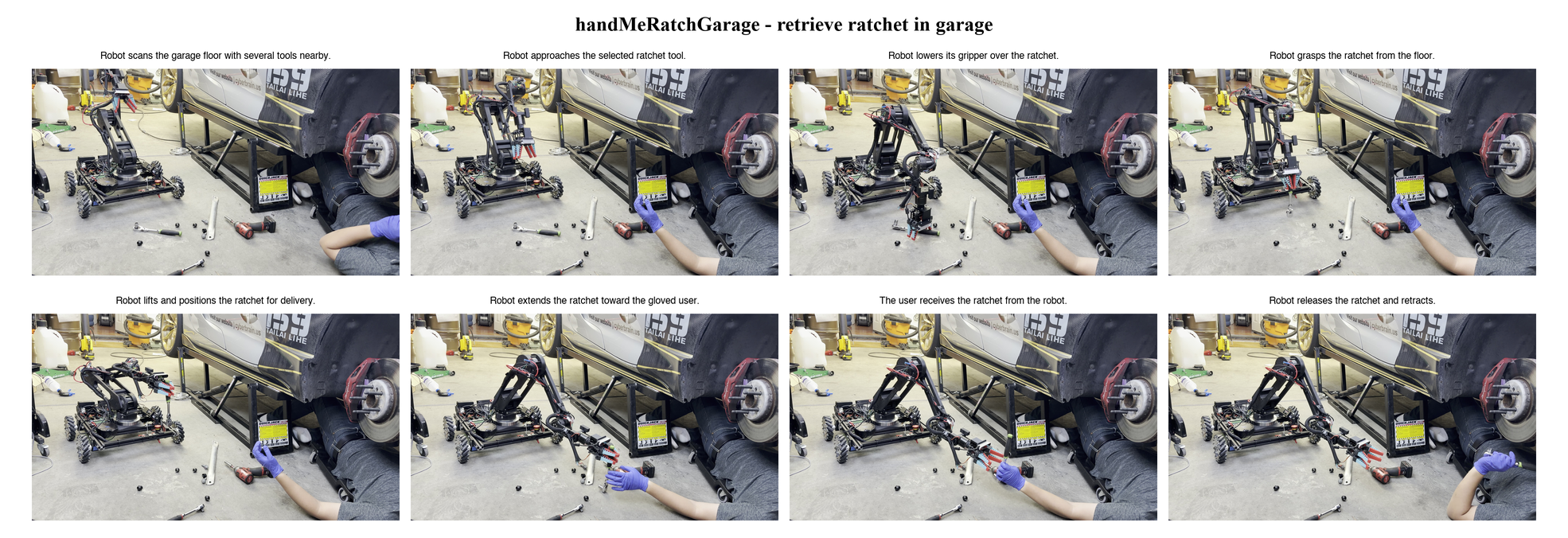}
  \caption{\textbf{``Hand me the green ratchet.''} Retrieving a tool in a garage setting.}
  \label{fig:demo-ratchet}
\end{figure}

\begin{figure}[H]
  \centering
  \includegraphics[width=\linewidth]{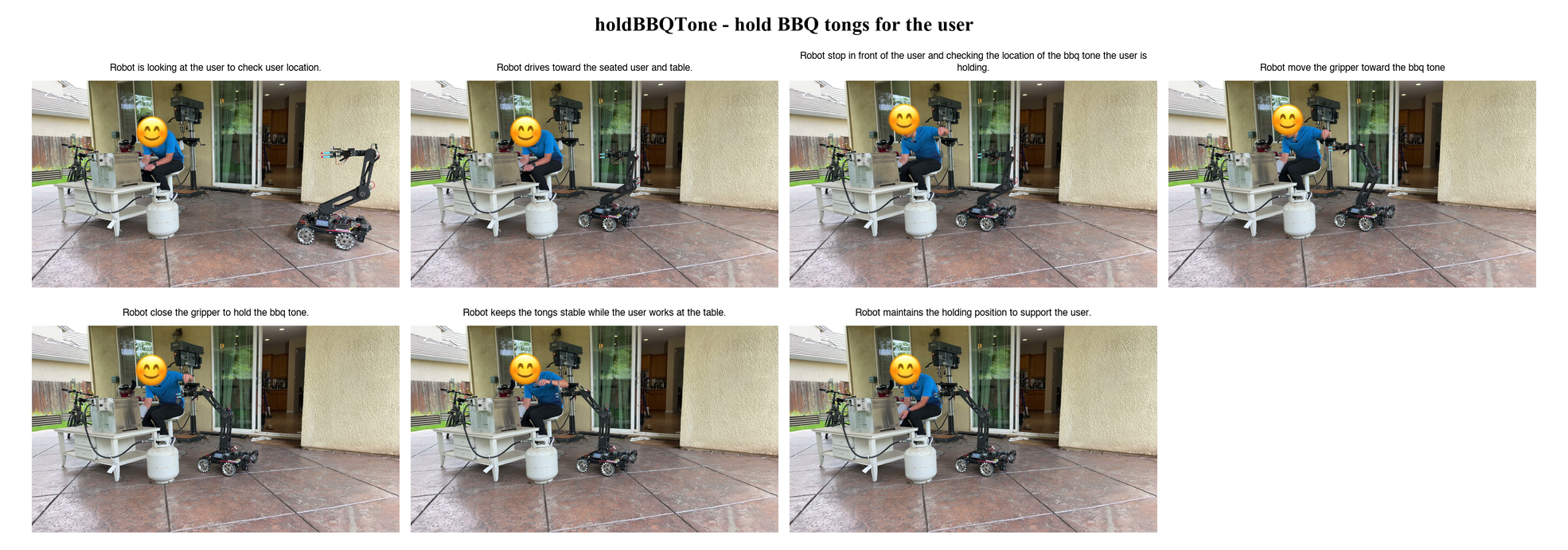}
  \caption{\textbf{``Hold the BBQ tongs.''} Grasping and presenting the tongs to the person.}
  \label{fig:demo-bbq}
\end{figure}

\begin{figure}[H]
  \centering
  \includegraphics[width=\linewidth]{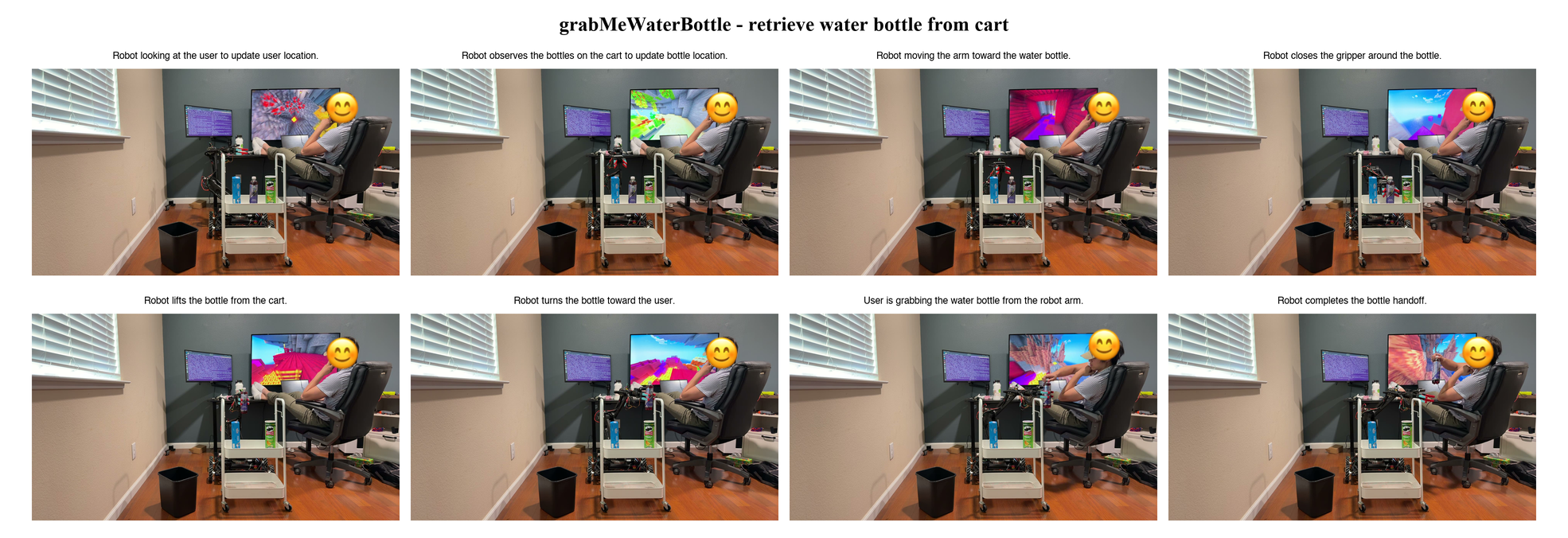}
  \caption{\textbf{``Grab me the water bottle.''} Object retrieval and handover.}
  \label{fig:demo-waterbottle}
\end{figure}

\begin{figure}[H]
  \centering
  \includegraphics[width=0.85\linewidth]{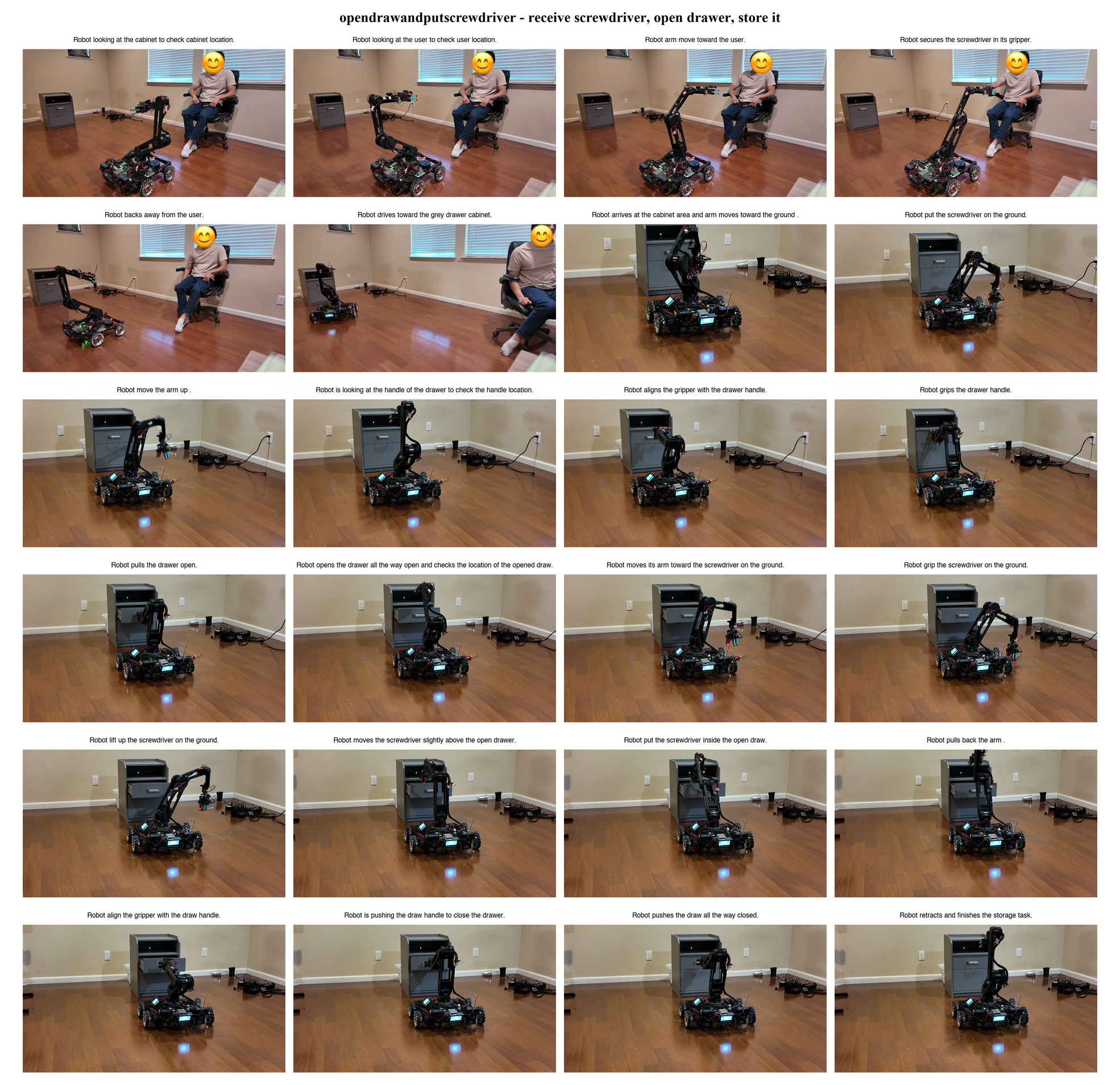}
  \caption{\textbf{``Take the screwdriver to the drawer.''} A two-stage manipulation:
    open the drawer, then place the screwdriver inside.}
  \label{fig:demo-screwdriver}
\end{figure}

\begin{figure}[H]
  \centering
  \includegraphics[width=\linewidth]{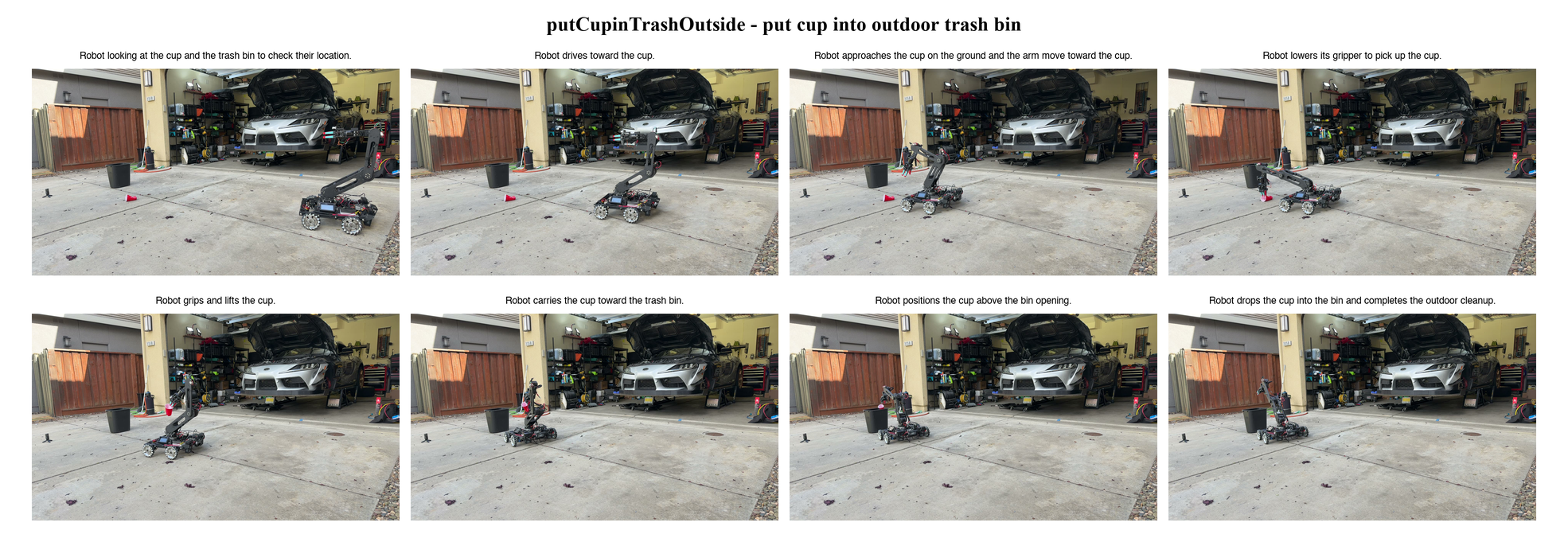}
  \caption{\textbf{``Put the cup in the trash can.''} (Outdoor) A multi-step task: pick up the cup, navigate outside, and dispose of it.}
  \label{fig:demo-trash}
\end{figure}

\begin{figure}[H]
  \centering
  \includegraphics[width=\linewidth]{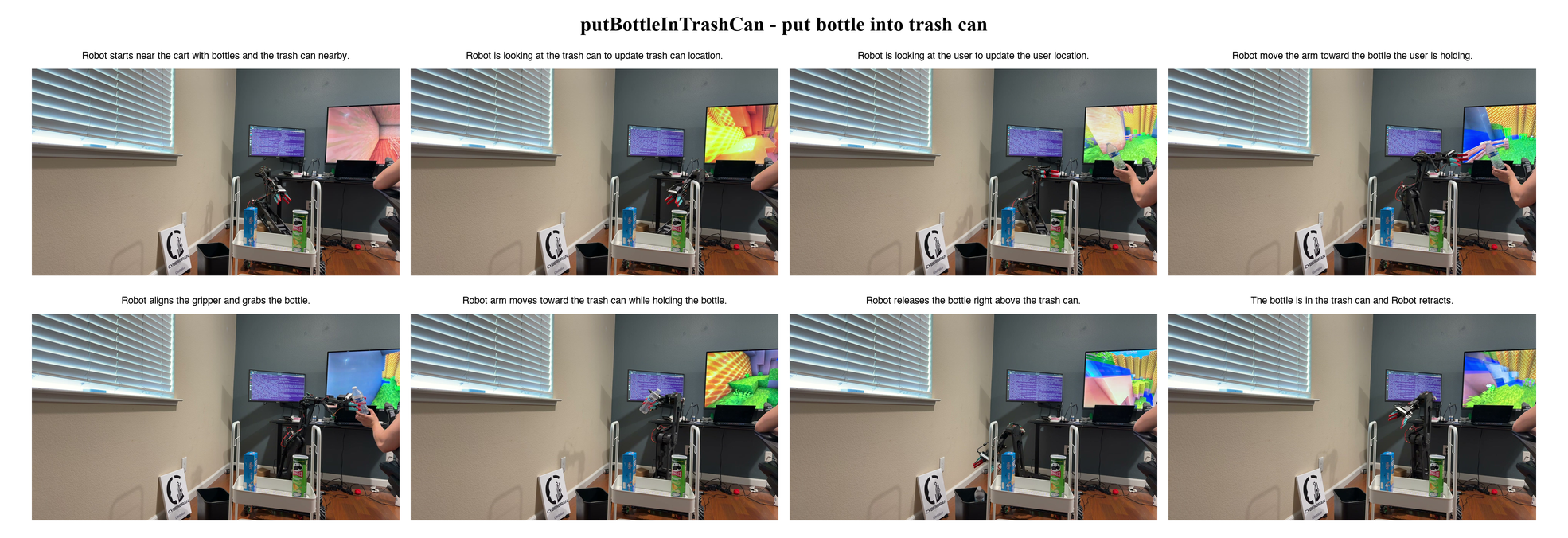}
  \caption{\textbf{``Put the bottle in the trash can.''} A multi-step task: pick up the bottle, locate the trash can, and dispose of it.}
  \label{fig:demo-bottletrash}
\end{figure}

Figures~\ref{fig:demo-popcorn}--\ref{fig:demo-bottletrash} show representative
frame sequences from the live demos summarized in
Table~\ref{tab:interaction}, each capturing the dialogue-driven handover from
the robot's first look at the scene through grasp and delivery to the person.

\section{Performance Comparison Results}
\label{app:benchmark}
\input{wcm_assets/tables/table1_benchmark.tex}

\section{Ablations}
\label{app:ablation}
\input{wcm_assets/tables/table3_ablation.tex}

%% file: wcm_assets/tables/table2_interaction.tex
\begin{table}[H]
\centering
\caption{\textbf{Human--robot interaction demos.} Per-task success rate (\%) of WCM
on nine live interaction tasks, each coupling dialogue with physical handover to a
person. Task labels are the spoken instructions of the matching demo figures
(Figs.~\ref{fig:demo-popcorn}--\ref{fig:demo-bottletrash}). Because conventional VLA
policies typically do not treat continuous dialogue, interpretable interruption, and
teaching mode as core closed-loop capabilities, we report WCM only.}
\label{tab:interaction}
\small
\setlength{\tabcolsep}{10pt}
\renewcommand{\arraystretch}{1.25}
\begin{tabular}{@{}l ccc@{}}
\toprule
\textbf{Spoken instruction} & \textbf{SR (\%)} & \textbf{Trials} & \textbf{Time (s)} \\
\midrule
``Grab me the popcorn.''                & 85 & 11/13 & 72  \\
``Hand me the honey.''                  & 79 & 11/14 & 34  \\
``Hand me the jacket.''                 & 67 & 6/9   & 54  \\
``Hand me the green ratchet.''          & 64 & 7/11  & 36  \\
``Hold the BBQ tongs.''                 & 86 & 12/14 & 48  \\
``Grab me the water bottle.''           & 83\textsuperscript{1} & 10/12 & 30  \\
``Take the screwdriver to the drawer.'' & 69\textsuperscript{2} & 9/13  & 165 \\
``Put the cup in the trash can.'' (Outdoor) & 67\textsuperscript{1} & 10/15 & 46  \\
``Put the bottle in the trash can.''    & 64\textsuperscript{1} & 9/14  & 34  \\
\bottomrule
\end{tabular}

\smallskip
\begin{minipage}{\linewidth}
\footnotesize \textsuperscript{1}\,Held out from the CoT training set; solved zero-shot by
transfer. \textsuperscript{2}\,Held out and acquired through teaching mode: the reported
$69\%$ is post-teaching but pre-distillation (below $20\%$ before teaching, $82\%$ after
CoT distillation). \textbf{Trials} reports successful\,/\,total attempts; \textbf{SR} is
their ratio. \textbf{Time} is a typical end-to-end run time (seconds), from robot-motion
onset to task completion; values are approximate.
\end{minipage}
\end{table}

%% file: wcm_assets/tables/table1_benchmark.tex
%
%
\begin{table}[H]
\centering
\caption{\textbf{Case-study comparison.} Per-task comparison of \textbf{WCM (Ours)}
against \textbf{X-Sq.} (X-Square Robot; model: \textbf{WALL-OSS}) on six tasks
across three metrics. Arrows give the better direction; \textbf{bold} marks the
better method per compared cell.}
\label{tab:benchmark}
\footnotesize
\setlength{\tabcolsep}{3pt}
\renewcommand{\arraystretch}{1.15}
\begin{tabular}{@{}l cc cc c@{}}
\toprule
& \multicolumn{2}{c}{Movement time (s)\,$\downarrow$}
& \multicolumn{2}{c}{GPU busy time (s)\,$\downarrow$}
& \multicolumn{1}{c}{Success (\%)\,$\uparrow$} \\
\cmidrule(lr){2-3}\cmidrule(lr){4-5}\cmidrule(lr){6-6}
Task & WCM & X-Sq. & WCM & X-Sq.\,(est.) & WCM \\
\midrule
ColorPlateSort    & \textbf{25.0} & 109.0 & \textbf{19.4} & 109.0 & 92 \\
LabelRecognition  & \textbf{36.0} & 116.0 & \textbf{18.4} & 116.0 & 90 \\
PickUpFruits      & \textbf{16.6} & 18.5  & \textbf{6.4}  & 18.5  & 90 \\
ThrowTrash        & \textbf{80.5} & 258.0 & \textbf{22.5} & 258.0 & 86 \\
PickUpCube        & \textbf{16.1} & 43.2  & \textbf{5.6}  & 43.2  & 92 \\
SolveMathProblem  & \textbf{57.0} & 76.0  & \textbf{59.9} & 76.0  & 78 \\
\cmidrule(lr){1-6}
\textbf{Average}  & \textbf{38.5} & 103.5 & \textbf{22.0} & 103.5 & 88.0 \\
\bottomrule
\end{tabular}

\smallskip
\begin{minipage}{\linewidth}
\footnotesize \textbf{Movement time} is measured from the onset of robot motion to
completion of the target task, matching the reference videos of the baseline (which start
when the robot begins to move). \textbf{GPU busy time} is the wall-clock time the GPU
spends computing for the task. A VLA runs on the GPU at every control step, so for
X-Square we estimate its GPU busy time as the full movement time (continuous occupancy);
WCM uses the GPU only intermittently, freeing it for other robots. WCM also spends
$3$--$5\,$s on a post-task summary, so its GPU stays busy slightly past the end of motion,
which is why GPU busy time can exceed movement time (e.g., \textsf{SolveMathProblem},
$59.9$ vs $57.0\,$s). \textbf{Success.} X-Square did not provide per-task success rates,
so the success column reports WCM only.
\end{minipage}
\end{table}


%% file: wcm_assets/tables/table3_ablation.tex
\begin{table}[H]
\centering
\caption{\textbf{Ablations.} Removing the \textbf{asynchronous runtime} (the robot must
halt to think between actions) inflates end-to-end run time; replacing \textbf{INTERACTION-AWARE SENSING LAYER }
with a plain VLM and 2D grounding (no part-level interaction cues) reduces the success
rate. Both are evaluated on the same five tasks. ``WCM'' is the full system---its times
and success rates match Table~\ref{tab:interaction}; arrows give the better direction and
\textbf{bold} marks the full system.}
\label{tab:ablation}
\small
\setlength{\tabcolsep}{6pt}
\renewcommand{\arraystretch}{1.25}
\begin{tabular}{@{}l cc cc@{}}
\toprule
& \multicolumn{2}{c}{Time (s)\,$\downarrow$}
& \multicolumn{2}{c}{Success (\%)\,$\uparrow$} \\
\cmidrule(lr){2-3}\cmidrule(lr){4-5}
Task & WCM & \makecell{w/o\\Async} & WCM & \makecell{w/o\\Sensing} \\
\midrule
``Hand me the green ratchet.''          & \textbf{36}  & 61  & \textbf{64} & 18 \\
``Grab me the water bottle.''           & \textbf{30}  & 47  & \textbf{83} & 27 \\
``Take the screwdriver to the drawer.'' & \textbf{165} & 254 & \textbf{69} & 0  \\
``Put the cup in the trash can.'' (Outdoor) & \textbf{46}  & 63  & \textbf{67} & 18 \\
``Put the bottle in the trash can.''    & \textbf{34}  & 49  & \textbf{64} & 9  \\
\bottomrule
\end{tabular}
\end{table}

%% file: paper.bbl
\begin{thebibliography}{37}
\providecommand{\natexlab}[1]{#1}
\providecommand{\url}[1]{\texttt{#1}}
\expandafter\ifx\csname urlstyle\endcsname\relax
  \providecommand{\doi}[1]{doi: #1}\else
  \providecommand{\doi}{doi: \begingroup \urlstyle{rm}\Url}\fi

\bibitem[Agarwal et~al.(2025)Agarwal, Ali, Bala, Balaji, Barker, Cai,
  et~al.]{cosmos}
Niket Agarwal, Arslan Ali, Maciej Bala, Yogesh Balaji, Erik Barker, Tiffany
  Cai, et~al.
\newblock Cosmos world foundation model platform for physical ai.
\newblock \emph{arXiv preprint arXiv:2501.03575}, 2025.

\bibitem[Bjorck et~al.(2025)Bjorck, Castaneda, Cherniadev, Da, Ding, Fan,
  et~al.]{grootn1}
Johan Bjorck, Fernando Castaneda, Nikita Cherniadev, Xingye Da, Runyu Ding,
  Linxi Fan, et~al.
\newblock {GR00T N1}: An open foundation model for generalist humanoid robots.
\newblock \emph{arXiv preprint arXiv:2503.14734}, 2025.

\bibitem[Black et~al.(2025{\natexlab{a}})Black, Brown, Darpinian, Dhabalia,
  Driess, Esmail, Equi, et~al.]{pi05}
Kevin Black, Noah Brown, James Darpinian, Karan Dhabalia, Danny Driess, Adnan
  Esmail, Michael~Robert Equi, et~al.
\newblock {$\pi_{0.5}$}: A vision-language-action model with open-world
  generalization.
\newblock In \emph{Proceedings of The 9th Conference on Robot Learning}, volume
  305 of \emph{Proceedings of Machine Learning Research}, pages 17--40. PMLR,
  2025{\natexlab{a}}.

\bibitem[Black et~al.(2025{\natexlab{b}})Black, Brown, Driess, et~al.]{pi0}
Kevin Black, Noah Brown, Danny Driess, et~al.
\newblock {$\pi_0$}: A vision-language-action flow model for general robot
  control.
\newblock In \emph{Robotics: Science and Systems (RSS)}, 2025{\natexlab{b}}.
\newblock arXiv:2410.24164.

\bibitem[Bousmalis et~al.(2023)Bousmalis, Vezzani, Rao, Devin, Lee, Bauza,
  Davchev, Zhou, et~al.]{robocat}
Konstantinos Bousmalis, Giulia Vezzani, Dushyant Rao, Coline Devin, Alex~X.
  Lee, Maria Bauza, Todor Davchev, Yuxiang Zhou, et~al.
\newblock Robocat: A self-improving generalist agent for robotic manipulation.
\newblock \emph{Transactions on Machine Learning Research}, 2023.
\newblock arXiv:2306.11706.

\bibitem[Brohan et~al.(2023)Brohan, Brown, Carbajal, et~al.]{rt1}
Anthony Brohan, Noah Brown, Justice Carbajal, et~al.
\newblock {RT-1}: Robotics transformer for real-world control at scale.
\newblock In \emph{Robotics: Science and Systems (RSS)}, 2023.
\newblock arXiv:2212.06817.

\bibitem[Bruce et~al.(2024)Bruce, Dennis, Edwards, et~al.]{genie}
Jake Bruce, Michael~D Dennis, Ashley Edwards, et~al.
\newblock Genie: Generative interactive environments.
\newblock In \emph{Proceedings of the 41st International Conference on Machine
  Learning}, volume 235 of \emph{Proceedings of Machine Learning Research},
  pages 4603--4623. PMLR, 2024.
\newblock arXiv:2402.15391.

\bibitem[Carion et~al.(2026)Carion, Gustafson, Hu, Debnath, Hu, Suris, Ryali,
  Alwala, Khedr, Huang, Lei, Ma, Guo, Kalla, Marks, Greer, Wang, Sun,
  R{\"a}dle, Afouras, Mavroudi, Xu, Wu, Zhou, Momeni, Ding, Vaze, Porcher, Li,
  Li, Kamath, Cheng, Doll{\'a}r, Ravi, Saenko, Zhang, and Feichtenhofer]{sam3}
Nicolas Carion, Laura Gustafson, Yuan-Ting Hu, Shoubhik Debnath, Ronghang Hu,
  Didac Suris, Chaitanya Ryali, Kalyan~Vasudev Alwala, Haitham Khedr, Andrew
  Huang, Jie Lei, Tengyu Ma, Baishan Guo, Arpit Kalla, Markus Marks, Joseph
  Greer, Meng Wang, Peize Sun, Roman R{\"a}dle, Triantafyllos Afouras,
  Effrosyni Mavroudi, Katherine Xu, Tsung-Han Wu, Yu~Zhou, Liliane Momeni,
  Shuangrui Ding, Sagar Vaze, Francois Porcher, Feng Li, Siyuan Li, Aishwarya
  Kamath, Ho~Kei Cheng, Piotr Doll{\'a}r, Nikhila Ravi, Kate Saenko, Pengchuan
  Zhang, and Christoph Feichtenhofer.
\newblock Sam 3: Segment anything with concepts.
\newblock In \emph{International Conference on Learning Representations
  (ICLR)}, 2026.
\newblock arXiv:2511.16719.

\bibitem[Chen et~al.(2025)Chen, Son, and Kusari]{matpredict}
Yuzhen Chen, Hojun Son, and Arpan Kusari.
\newblock Matpredict: A dataset and benchmark for learning material properties
  of diverse indoor objects.
\newblock \emph{arXiv preprint arXiv:2505.13201}, 2025.

\bibitem[Clark and Brennan(1991)]{clarkbrennan}
Herbert~H. Clark and Susan~E. Brennan.
\newblock Grounding in communication.
\newblock In Lauren~B. Resnick, John~M. Levine, and Stephanie~D. Teasley,
  editors, \emph{Perspectives on Socially Shared Cognition}, pages 127--149.
  American Psychological Association, 1991.

\bibitem[Deng et~al.(2023)Deng, Prasad, Fernandez, et~al.]{deng2023implicit}
Yuntian Deng, Kiran Prasad, Roland Fernandez, et~al.
\newblock Implicit chain of thought reasoning via knowledge distillation.
\newblock \emph{arXiv preprint arXiv:2311.01460}, 2023.

\bibitem[Deng et~al.(2024)Deng, Choi, and Shieber]{deng2024explicit}
Yuntian Deng, Yejin Choi, and Stuart Shieber.
\newblock From explicit cot to implicit cot: Learning to internalize cot step
  by step.
\newblock \emph{arXiv preprint arXiv:2405.14838}, 2024.

\bibitem[Driess et~al.(2023)Driess, Xia, Sajjadi, et~al.]{palme}
Danny Driess, Fei Xia, Mehdi S.~M. Sajjadi, et~al.
\newblock {PaLM-E}: An embodied multimodal language model.
\newblock In \emph{Proceedings of the 40th International Conference on Machine
  Learning}, volume 202 of \emph{Proceedings of Machine Learning Research},
  pages 8469--8488. PMLR, 2023.
\newblock arXiv:2303.03378.

\bibitem[Ewen et~al.(2024)Ewen, Chen, Chen, Li, Bagali, Gunjal, and
  Vasudevan]{ewen2024feel}
Parker Ewen, Hao Chen, Yuzhen Chen, Anran Li, Anup Bagali, Gitesh Gunjal, and
  Ram Vasudevan.
\newblock You've got to feel it to believe it: Multi-modal bayesian inference
  for semantic and property prediction.
\newblock In \emph{Robotics: Science and Systems (RSS)}, 2024.
\newblock arXiv:2402.05872.

\bibitem[Ghosh et~al.(2024)Ghosh, Walke, Pertsch, Black, Mees, Dasari,
  et~al.]{octo}
Dibya Ghosh, Homer~Rich Walke, Karl Pertsch, Kevin Black, Oier Mees, Sudeep
  Dasari, et~al.
\newblock Octo: An open-source generalist robot policy.
\newblock In \emph{Robotics: Science and Systems (RSS)}, 2024.
\newblock arXiv:2405.12213.

\bibitem[Ha and Schmidhuber(2018)]{worldmodels}
David Ha and J{\"u}rgen Schmidhuber.
\newblock World models.
\newblock \emph{arXiv preprint arXiv:1803.10122}, 2018.

\bibitem[Horvitz(1999)]{horvitz}
Eric Horvitz.
\newblock Principles of mixed-initiative user interfaces.
\newblock In \emph{Proceedings of the SIGCHI Conference on Human Factors in
  Computing Systems (CHI)}, pages 159--166, 1999.

\bibitem[Hsieh et~al.(2023)Hsieh, Li, Yeh, et~al.]{distill}
Cheng-Yu Hsieh, Chun-Liang Li, Chih-Kuan Yeh, et~al.
\newblock Distilling step-by-step! outperforming larger language models with
  less training data and smaller model sizes.
\newblock In \emph{Findings of the Association for Computational Linguistics
  (ACL)}, pages 8003--8017, 2023.
\newblock arXiv:2305.02301.

\bibitem[Huang et~al.(2023)Huang, Xia, Xiao, et~al.]{huang2022inner}
Wenlong Huang, Fei Xia, Ted Xiao, et~al.
\newblock Inner monologue: Embodied reasoning through planning with language
  models.
\newblock In \emph{Proceedings of The 6th Conference on Robot Learning}, volume
  205 of \emph{Proceedings of Machine Learning Research}, pages 1769--1782.
  PMLR, 2023.
\newblock arXiv:2207.05608.

\bibitem[Ichter et~al.(2023)Ichter, Brohan, Chebotar, Finn, Hausman, Herzog,
  Ho, et~al.]{saycan}
Brian Ichter, Anthony Brohan, Yevgen Chebotar, Chelsea Finn, Karol Hausman,
  Alexander Herzog, Daniel Ho, et~al.
\newblock Do as i can, not as i say: Grounding language in robotic affordances.
\newblock In \emph{Proceedings of The 6th Conference on Robot Learning}, volume
  205 of \emph{Proceedings of Machine Learning Research}, pages 287--318. PMLR,
  2023.
\newblock arXiv:2204.01691.

\bibitem[Jang et~al.(2025)Jang, Ye, Lin, Xiang, Bjorck, Fang, Hu,
  et~al.]{dreamgen}
Joel Jang, Seonghyeon Ye, Zongyu Lin, Jiannan Xiang, Johan Bjorck, Yu~Fang,
  Fengyuan Hu, et~al.
\newblock Dreamgen: Unlocking generalization in robot learning through video
  world models.
\newblock In \emph{Proceedings of The 9th Conference on Robot Learning}, volume
  305 of \emph{Proceedings of Machine Learning Research}, pages 5170--5194.
  PMLR, 2025.

\bibitem[Kim et~al.(2025)Kim, Pertsch, Karamcheti, et~al.]{openvla}
Moo~Jin Kim, Karl Pertsch, Siddharth Karamcheti, et~al.
\newblock {OpenVLA}: An open-source vision-language-action model.
\newblock In \emph{Proceedings of The 8th Conference on Robot Learning}, volume
  270 of \emph{Proceedings of Machine Learning Research}, pages 2679--2713.
  PMLR, 2025.
\newblock arXiv:2406.09246.

\bibitem[Liang et~al.(2023)Liang, Huang, Xia, et~al.]{liang2023code}
Jacky Liang, Wenlong Huang, Fei Xia, et~al.
\newblock Code as policies: Language model programs for embodied control.
\newblock In \emph{IEEE International Conference on Robotics and Automation
  (ICRA)}, pages 9493--9500, 2023.
\newblock arXiv:2209.07753.

\bibitem[Lightman et~al.(2024)Lightman, Kosaraju, Burda,
  et~al.]{lightman2024verify}
Hunter Lightman, Vineet Kosaraju, Yura Burda, et~al.
\newblock Let's verify step by step.
\newblock In \emph{International Conference on Learning Representations
  (ICLR)}, 2024.
\newblock arXiv:2305.20050.

\bibitem[Liu et~al.(2024)Liu, Zeng, Ren, Li, Zhang, Yang, Jiang, Li, Yang, Su,
  Zhu, and Zhang]{groundingdino}
Shilong Liu, Zhaoyang Zeng, Tianhe Ren, Feng Li, Hao Zhang, Jie Yang, Qing
  Jiang, Chunyuan Li, Jianwei Yang, Hang Su, Jun Zhu, and Lei Zhang.
\newblock Grounding dino: Marrying dino with grounded pre-training for open-set
  object detection.
\newblock In \emph{European Conference on Computer Vision (ECCV)}, pages
  38--55, 2024.
\newblock arXiv:2303.05499.

\bibitem[Nye et~al.(2021)Nye, Andreassen, Gur-Ari, et~al.]{nye2021show}
Maxwell Nye, Anders Andreassen, Guy Gur-Ari, et~al.
\newblock Show your work: Scratchpads for intermediate computation with
  language models.
\newblock \emph{arXiv preprint arXiv:2112.00114}, 2021.

\bibitem[Ren et~al.(2024)Ren, Liu, Zeng, Lin, Li, Cao, Chen, Huang, Chen, Yan,
  Zeng, Zhang, Li, Yang, Li, Jiang, and Zhang]{groundedsam}
Tianhe Ren, Shilong Liu, Ailing Zeng, Jing Lin, Kunchang Li, He~Cao, Jiayu
  Chen, Xinyu Huang, Yukang Chen, Feng Yan, Zhaoyang Zeng, Hao Zhang, Feng Li,
  Jie Yang, Hongyang Li, Qing Jiang, and Lei Zhang.
\newblock Grounded sam: Assembling open-world models for diverse visual tasks.
\newblock \emph{arXiv preprint arXiv:2401.14159}, 2024.

\bibitem[Sutton et~al.(1999)Sutton, Precup, and Singh]{sutton1999between}
Richard~S. Sutton, Doina Precup, and Satinder Singh.
\newblock Between {MDP}s and semi-{MDP}s: A framework for temporal abstraction
  in reinforcement learning.
\newblock \emph{Artificial Intelligence}, 112\penalty0 (1--2):\penalty0
  181--211, 1999.

\bibitem[Uesato et~al.(2022)Uesato, Kushman, Kumar, et~al.]{uesato2022solving}
Jonathan Uesato, Nate Kushman, Ramana Kumar, et~al.
\newblock Solving math word problems with process- and outcome-based feedback.
\newblock \emph{arXiv preprint arXiv:2211.14275}, 2022.

\bibitem[Wang et~al.(2024)Wang, Xian, Chen, Wang, Wang, Fragkiadaki, Erickson,
  Held, and Gan]{robogen}
Yufei Wang, Zhou Xian, Feng Chen, Tsun-Hsuan Wang, Yian Wang, Katerina
  Fragkiadaki, Zackory Erickson, David Held, and Chuang Gan.
\newblock {RoboGen}: Towards unleashing infinite data for automated robot
  learning via generative simulation.
\newblock In \emph{Proceedings of the 41st International Conference on Machine
  Learning}, volume 235 of \emph{Proceedings of Machine Learning Research},
  pages 51936--51983. PMLR, 2024.

\bibitem[Wei et~al.(2022)Wei, Wang, Schuurmans, et~al.]{cot}
Jason Wei, Xuezhi Wang, Dale Schuurmans, et~al.
\newblock Chain-of-thought prompting elicits reasoning in large language
  models.
\newblock In \emph{Advances in Neural Information Processing Systems
  (NeurIPS)}, volume~35, pages 24824--24837, 2022.
\newblock arXiv:2201.11903.

\bibitem[Wu et~al.(2023)Wu, Escontrela, Hafner, Abbeel, and
  Goldberg]{daydreamer}
Philipp Wu, Alejandro Escontrela, Danijar Hafner, Pieter Abbeel, and Ken
  Goldberg.
\newblock Daydreamer: World models for physical robot learning.
\newblock In \emph{Proceedings of The 6th Conference on Robot Learning}, volume
  205 of \emph{Proceedings of Machine Learning Research}, pages 2226--2240.
  PMLR, 2023.

\bibitem[Zhou et~al.(2026{\natexlab{a}})Zhou, Bai, Chang, Wang, Liang, and
  Zhan]{stream3d}
Kaichen Zhou, Zeyang Bai, Xinhai Chang, Mengyu Wang, Paul~Pu Liang, and
  Fangneng Zhan.
\newblock Stream3d: Sequential multi-view 3d generation via evidential memory.
\newblock \emph{arXiv preprint arXiv:2605.21472}, 2026{\natexlab{a}}.

\bibitem[Zhou et~al.(2026{\natexlab{b}})Zhou, Chen, Zhan, Hua, Chen, Chang, Qu,
  Du, Liu, Liang, et~al.]{zhou2026gem}
Kaichen Zhou, Yuzhen Chen, Fangneng Zhan, Hang Hua, Grace Chen, Xinhai Chang,
  Ao~Qu, Yilun Du, Zhuang Liu, Paul~Pu Liang, et~al.
\newblock Gem-4d: Geometry-enhanced video world models for robot manipulation.
\newblock \emph{arXiv preprint arXiv:2605.22882}, 2026{\natexlab{b}}.

\bibitem[Zhou et~al.(2026{\natexlab{c}})Zhou, Wang, Chen, Beaudouin, Zhan,
  Liang, and Wang]{page4d}
Kaichen Zhou, Yuhan Wang, Grace Chen, Gaspard Beaudouin, Fangneng Zhan, Paul~Pu
  Liang, and Mengyu Wang.
\newblock Page-4d: Disentangled pose and geometry estimation for vggt-4d
  perception.
\newblock In \emph{International Conference on Learning Representations
  (ICLR)}, 2026{\natexlab{c}}.
\newblock arXiv:2510.17568.

\bibitem[Zhou et~al.(2024)Zhou, Du, Chen, Li, Yeung, and Gan]{robodreamer}
Siyuan Zhou, Yilun Du, Jiaben Chen, Yandong Li, Dit-Yan Yeung, and Chuang Gan.
\newblock Robodreamer: Learning compositional world models for robot
  imagination.
\newblock In \emph{Proceedings of the 41st International Conference on Machine
  Learning}, volume 235 of \emph{Proceedings of Machine Learning Research},
  pages 61885--61896. PMLR, 2024.

\bibitem[Zitkovich et~al.(2023)Zitkovich, Yu, Xu, Xu, Xiao, Xia, et~al.]{rt2}
Brianna Zitkovich, Tianhe Yu, Sichun Xu, Peng Xu, Ted Xiao, Fei Xia, et~al.
\newblock {RT-2}: Vision-language-action models transfer web knowledge to
  robotic control.
\newblock In \emph{Proceedings of The 7th Conference on Robot Learning}, volume
  229 of \emph{Proceedings of Machine Learning Research}, pages 2165--2183.
  PMLR, 2023.
\newblock arXiv:2307.15818.

\end{thebibliography}
